\documentclass[letterpaper, 10 pt, conference]{ieeeconf}  
\usepackage{graphicx}
\usepackage{amsmath}
\usepackage{amssymb}
\usepackage{placeins}
\usepackage{subcaption} 
\usepackage{caption}
\usepackage{xcolor}
\usepackage{hyperref}
\hypersetup{
    colorlinks = true,
    linkcolor = blue,
    urlcolor = blue,
}

\usepackage{booktabs}
\usepackage{cite}
\let\labelindent\relax
\usepackage{enumitem}
\usepackage{multirow}
\usepackage{tabularx}
\usepackage{algorithm}
\usepackage{algorithmicx}
\usepackage{algpseudocode}
\usepackage{url}
\usepackage{siunitx}

\newcommand{\plus}{\thinspace\raisebox{0pt}[0pt][0pt]{+}\thinspace}
\newcommand{\model}{EnduRL} 
\newcommand{\nop}[1]{} 
\newcommand{\round}[1]{\ensuremath{\lfloor#1\rceil}}

\IEEEoverridecommandlockouts                              
\title{\LARGE \bf
\model: Enhancing Safety, Stability, and Efficiency of Mixed Traffic Under Real-World Perturbations Via Reinforcement Learning    
}
\author{Bibek Poudel$^{1}$, Weizi Li$^{1}$, Kevin Heaslip$^{2}$ 
\thanks{$^{1}$Bibek Poudel and Weizi Li are with Min H. Kao Department of Electrical Engineering and Computer Science at University of Tennessee, Knoxville, TN, USA {\tt\small bpoudel3@vols.utk.edu, weizili@utk.edu}}%
\thanks{$^{2}$Kevin Heaslip is with Department of Civil and Environmental Engineering at University of Tennessee, Knoxville, TN, USA {\tt\small kheaslip@utk.edu}}
}

\begin{document}

\maketitle
\thispagestyle{empty}
\pagestyle{empty}


\begin{abstract}
Human-driven vehicles (HVs) amplify naturally occurring perturbations in traffic, leading to congestion -- a major contributor to increased fuel consumption, higher collision risks, and reduced road capacity utilization. While previous research demonstrates that Robot Vehicles (RVs) can be leveraged to mitigate these issues, most such studies rely on simulations with simplistic models of human car-following behaviors. 
In this work, we analyze real-world driving trajectories and extract a wide range of acceleration profiles. We then incorporates these profiles into simulations for training RVs to mitigate congestion. 
We evaluate the safety, efficiency, and stability of mixed traffic via comprehensive experiments conducted in two mixed traffic environments (Ring and Bottleneck) at various traffic densities, configurations, and RV penetration rates. 
The results show that under real-world perturbations, prior RV controllers experience performance degradation on all three objectives (sometimes even lower than $100\%$ HVs). 
To address this, we introduce a reinforcement learning based RV that employs a congestion stage classifier to optimize the safety, efficiency, and stability of mixed traffic. 
Our RVs demonstrate significant improvements: safety by up to $66\%$, efficiency by up to $54\%$, and stability by up to $97\%$.
\end{abstract}

\section{Introduction}
Every major city in the world faces traffic congestion, from infrastructure-induced bottlenecks 
on bridges to more subtle phenomena like phantom jams on highways. Unsteady traffic flow during congestion increases travel time, energy consumption, and collision risk~\cite{buchanan2015traffic}. 
While the causes of congestion are not fully comprehended, a prominent explanation attributing to congestion is the asymmetric driving theory~\cite{yeo2008asymmetric}. This theory posits that human driving is characterized by frequent under- and over-reactions, stemming from heterogeneous driving styles and individual differences in estimation, reaction, and actuation times.
These factors will intensify in dense traffic and eventually lead to congestion~\cite{yeo2009understanding}. 

As more vehicles with varying levels of autonomy are introduced into our transportation system, the concept of \emph{mixed traffic control}, i.e., using Robot Vehicles (RVs) to address perturbations produced and amplified by Human-driven Vehicles (HVs), has emerged~\cite{di2021survey,Villarreal2023Pixel,Villarreal2024Eco,Villarreal2023Can}. Various control strategies are introduced such as those based on heuristics~\cite{stern2018dissipation}, models of longitudinal dynamics~\cite{horn2013suppressing, rajamani2011vehicle}, and machine learning~\cite{wu2021flow}. These techniques have been proven effective in reducing congestion in scenarios such as intersections~\cite{wang2023learning}, highways~\cite{yildirim2022prediction}, bottlenecks~\cite{vinitsky2018lagrangian}, and large networks~\cite{Wang2024Privacy} at RV penetration rates as low as $5\%$. 
Most of these studies are carried out in simulation using software such as Simulation of Urban MObility (SUMO)~\cite{lopez2018microscopic}, where model-based methods dominate in representing car-following behaviors of HVs~\cite{treiber2013traffic, treiber2000congested}.



Among various car-following models for enabling longitudinal control of a vehicle, Intelligent Driver Model (IDM)~\cite{treiber2017intelligent} is a popular choice with minimal parameters. 
To simulate heterogeneous driving behaviors that account for drivers' individual differences, stochasticity is often integrated with the model, and imperfections in perception, processing, and actuation are accounted for in the simulation~\cite{sumoDriverState}. 
However, even with the adjustments, car-following behaviors of HVs are largely constrained: $92\%$ of the accelerations derived using the stochastic IDM model with the default parameters~\cite{treiber2013traffic} lie within the range $[-0.5,0.5]~m/s^2$. In contrast, only $68\%$ of real-world accelerations (extracted from the I-$24$ MOTION dataset~\cite{gloudemans202324}) fall within the same range, resulting in a $\mathbf{24\%}$ \textbf{difference}. Further, real-world accelerations have \emph{a long tail} that extends up to $[-3,3]~m/s^2$. 
This observation reveals that simulated car-following mainly depicts safe or timid behaviors, lacking the representation of more aggressive real-world behaviors such as sudden braking and rapid accelerations~\cite{higgs2013two,peng2016new}. 
Subsequently, \emph{RVs developed and validated in improving traffic conditions with only timid driving behaviors may struggle to handle real-world perturbations}, thereby impeding their effectiveness in achieving the goals of safety, stability, and efficiency

Limited research to date bridges the gap between IDM and real-world driving~\cite{albeaik2022limitations}. Enhancements to IDM, including the integration of random noise and calibration with real-world driving data, are proposed to improve simulation accuracy. However, they fall short in capturing the full traffic variability and often lack broad applicability due to restrictive assumptions and artificial constraints on vehicle behaviors~\cite{wu2017flow, kreidieh2018dissipating, das2021saint, chou2022lord, vinitsky2018benchmarks, sridhar2021piecewise}. 
We introduce \model{}, a framework overcomes these limitations by periodically sampling accelerations from the real-world dataset during car-following. 
This produces increased variability in driving behavior with more aggressive accelerations and decelerations. 
In addition, we propose a reinforcement learning (RL)-based RV that leverages downstream traffic information to forecast congestion stages and takes preemptive actions to improve traffic conditions.

We compare \model{} with RVs developed in prior studies through comprehensive experiments in two mixed traffic environments: Ring~\cite{sugiyama2008traffic} and Bottleneck~\cite{vinitsky2018lagrangian} (see Fig.~\ref{fig:systematic}) under $1500+$ simulation runs. To evaluate safety, we use two surrogate  measures, time to collision (TTC) and deceleration rate to avoid a crash (DRAC); for efficiency we measure fuel economy and throughput; for stability, we measure acceleration variation and wave attenuation. 
These metrics are widely adopted in traffic engineering~\cite{gettman2003surrogate, tak2016study, rajamani2011vehicle}.
Our results show that in Ring, our RV improves both safety and efficiency by up to $\mathbf{54}\%$, and stability by up to $\mathbf{97}\%$. 
Whereas in Bottleneck, our RV can improves safety by up to $\mathbf{66}\%$, efficiency up to $\mathbf{41}\%$, and stability up to $\mathbf{34}\%$.  
To the best of our knowledge, \model{} is among the few studies that address the crucial gap between car-following behaviors in simulation and the real world and significantly improves the safety, stability, and efficiency of mixed traffic compared to previous studies. The project code can be found in the repository: \underline{\url{https://github.com/poudel-bibek/EnduRL}}

\section{Related Work}

To improve simulation fidelity, enhancements to IDM such as incorporating random noise~\cite{treiber2017intelligent} and calibrating IDM using real-world trajectories~\cite{kesting2008calibrating,li2016global}
are introduced. 
However, they are limited in reproducing the variability found in real-world traffic~\cite{sharath2020enhanced} and have limited generalizability~\cite{zhu2018modeling}.
Prior studies also impose artificial bounds that limit the acceleration range of HVs~\cite{wu2017flow, kreidieh2018dissipating, das2021saint, chou2022lord, vinitsky2018benchmarks, sridhar2021piecewise}. More recent studies have adopted machine learning for approximating human driving behaviors. These techniques include applying supervised learning to features extracted from real-world driving data, replacing IDM with deep neural networks~\cite{wang2017capturing}, and adopting deep RL with reward function tuned based on real-world data~\cite{zhu2020safe, zhu2018human}. Other studies propose equipping HVs with bilateral information, i.e., from both leader and follower vehicles to improve the traffic condition~\cite{shi2022bilateral}.
The key difference of our work to most existing studies is that we impose no artificial approximation or bounds. 
Instead, we directly extract and sample car-following dynamics 
from a real-world traffic dataset~\cite{gloudemans202324}.

Regarding traffic control, numerous RVs (as controllers) have been proposed~\cite{wu2017flow, kreidieh2018dissipating, das2021saint} and benchmarked~\cite{chou2022lord, vinitsky2018benchmarks, sridhar2021piecewise}. 
We broadly classify them into model-based, heuristic-based, and learning-based RVs. We test two model-based RVs, Bilateral Control Module (BCM)~\cite{horn2013suppressing} and Linear Adaptive Cruise Control (LACC). BCM is a linear acceleration model that uses relative speed and headway information from both the leader and follower vehicles, whereas LACC considers only the vehicle in front. 
For LACC, we use the constant time headway model described by Rajamani~\cite{rajamani2011vehicle} with the calibration parameters chosen to mimic the speed of IDM vehicles at equilibrium.  
Moreover, we evaluate two heuristic-based RVs, FollowerStopper (FS)~\cite{stern2018dissipation} and Proportional-integral with saturation (PIwS)~\cite{stern2018dissipation}. FS is a velocity controller that tracks a set average velocity and travels slightly below that velocity to open up a gap, allowing the RV to dampen oscillations and brake smoothly when needed. PIwS estimates the average equilibrium velocity of the vehicles in the network and then drives at the estimated velocity. Both FS and PIwS require a calibration on desired (or equilibrium) velocity and may fail to stabilize traffic if the equilibrium velocity is set high. 

\section{Methodology}
\label{sec:methodology}


We introduce Intelligent Driver Model and detail the components of \model{}.  
\vspace{-.5em}

\subsection{Intelligent Driver Model (IDM)}
IDM~\cite{treiber2000congested} assumes that drivers aim to maintain a safe distance from their leader while trying to reach their desired speed. 
IDM vehicles accelerate when the headway to the leader is large, and decelerate below a set maximum deceleration. The acceleration is given by: 
$$
    a_{IDM} = a\left[1-\left(\frac{v}{v_0}\right)^\delta - \left(\frac{s^*(v, \nabla v)}{s}\right)^2\right],
    \label{eq:idm1}
$$
\noindent where $s^*$ is the desired headway:
$$
    s^*(v, \nabla v) = s_0 + max(0, v\cdot T + \frac{v\nabla v}{2\sqrt{ab}}),
    \label{eq:idm2}
$$
where $a$ is the maximum acceleration, $v$ is the velocity, $v_0$ is the desired velocity, $\delta$ is the acceleration exponent, $s$ is the headway, $\nabla v$ is relative velocity to the leader vehicle, $s_0$ is the minimum gap, $T$ is the desired time headway, and $b$ is maximum deceleration. The parameter values are set according to Treiber and Kesting~\cite{treiber2013traffic}
as $a=1$, $b=1.5$, $T=1$, $\delta = 4$, and $s_0 = 2$. 
\vspace{-.5em}

\subsection{Sampling Real-world Perturbations}

We apply a car-following-filter~\cite{zhu2018modeling} to the I-$24$ MOTION~\cite{gloudemans202324} dataset and extract $172,000$ instantaneous accelerations. Analyzing the frequency and duration of these accelerations reveals a negative correlation. 
To ensure that these real-world perturbations are accurately mirrored in the simulation, we adopt a probabilistic approach. 
First, we uniformly sample the frequency of the perturbations within the observed range of $[10,~30]$ per HV, for every $6$ minutes of car-following. 
Then, we sample acceleration intensity uniformly within $[-3,~3]~m/s^2$.
For each selected intensity $A_i$, we find the most common duration $\hat{T}_{A_i}$ by linearly mapping $A_i$ within the minimum ($T_{\text{min}}$) and maximum ($T_{\text{max}}$) observed durations ($T_{\text{range}} = T_{\text{max}}-T_{\text{min}}$). 
We then sample $T_{A_i}$, the duration of $A_i$, from a piecewise triangular distribution using the following conditional probability density function: 
$$
P(T_{A_i} | \hat{T}_{A_i}) = 
\left\{
\begin{array}{ll}
\frac{2(T_{A_i} - T_{\text{min}})}{T_{\text{range}}(\hat{T}_{A_i} - T_{\text{min}})}, & T_{\text{min}} \leq T_{A_i} < \hat{T}_{A_i}, \\
\frac{2(T_{\text{max}} - T_{A_i})}{T_{\text{range}}(T_{\text{max}} - \hat{T}_{A_i})}, & \hat{T}_{A_i} \leq T_{A_i} \leq T_{\text{max}}.
\end{array}
\right.
$$
Lastly, we randomly assign each sampled acceleration to a HV during the testing of RVs. 




\subsection{Reinforcement Learning with Congestion Stage Classifier}
\label{subsec:leveraging}
\begin{figure}[t!]
		\centering
		\includegraphics[width=0.95\linewidth]{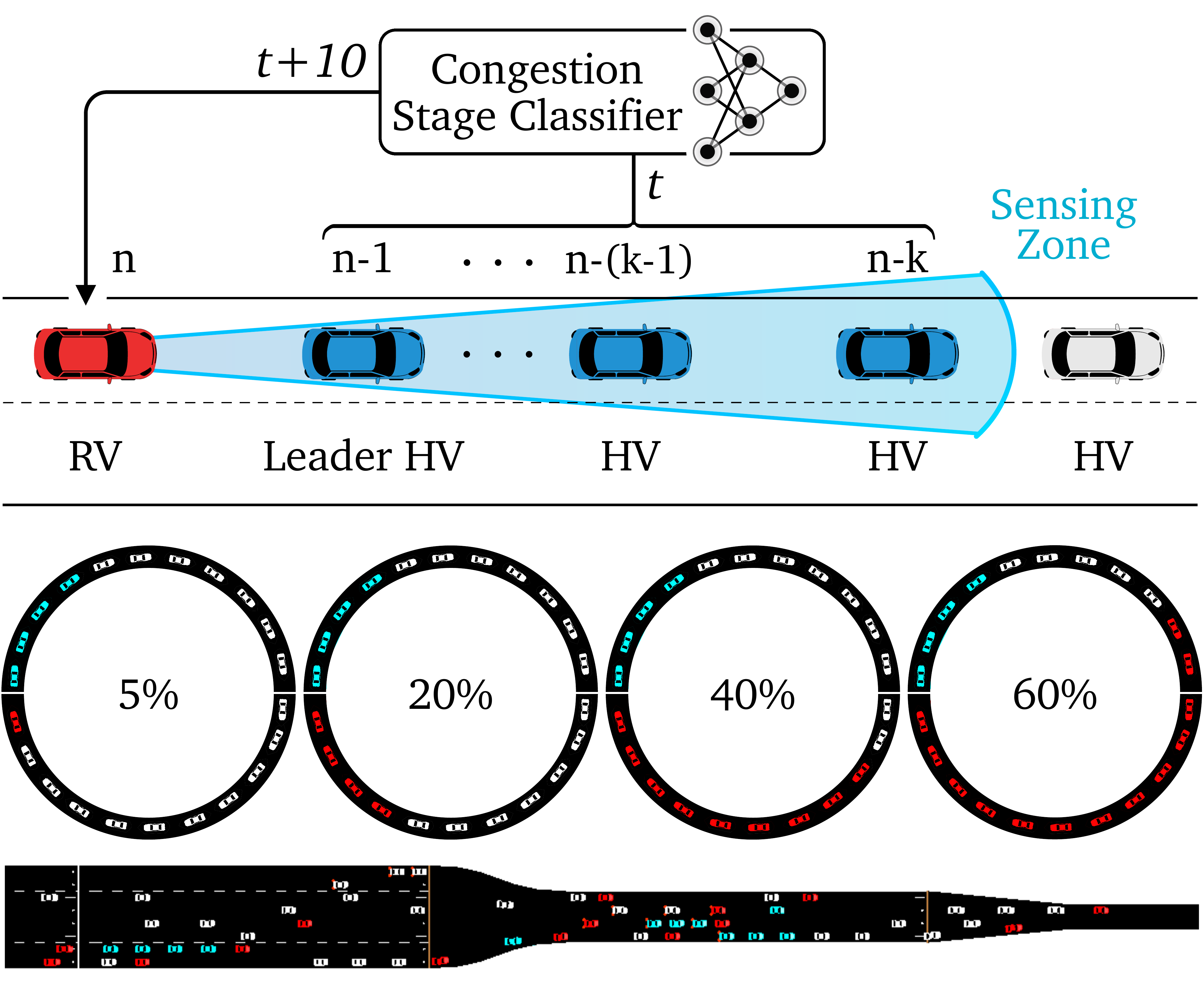}
		\caption{\small{TOP: The congestion stage classifier takes the position and velocity of the leader HVs of RV in the sensing zone to predict the congestion stage $10$ timesteps into the future, enabling pro-active responses of the RV. MIDDLE: Our RVs deployed at various penetration rates in Ring. When penetration rates $> 5\%$, RVs are arranged as a platoon. BOTTOM: Our RVs deployed at $40$\% penetration rate in Bottleneck (truncated version shown).}}
		\label{fig:systematic}
    \vspace{-1.5em}
\end{figure}

Our RL-based approach leverages Congestion Stage Classifier (CSC), a neural network trained with supervised learning on position and velocity from preceding cars inside a sensing zone (Fig.~\ref{fig:systematic}). The data is collected at various densities and categorized into six classes, namely `Forming', `Leaving', `Congested', `Free flow', `Undefined', and `No Vehicle', 
inspired by asymmetric driving theory~\cite{yeo2008asymmetric}. 
According to the theory, human drivers underestimate the required space headway during deceleration: when congestion is forming, availability of space headway decreases from one vehicle to the next as we move downstream within the sensing zone. Hence, a monotonic decrease in space headway is labeled congestion `Forming.' 
The theory also suggests that human drivers overestimate the space headway during acceleration: when a congestion is relieved, availability of space headway increases from one vehicle to the next as we move downstream within the sensing zone. Such a monotonic increase in space headway is labeled congestion `Leaving.'
When all distances are above a threshold without a clear pattern, `Free-Flow' is labeled; if all distances fall below a threshold, `Congested' is labeled. 
With this labeling scheme, CSC predicts and classifies the congestion stage $10$ timesteps in advance. 
We then incorporate the predictions by CSC into the observations and reward function of the RV.  
One CSC is trained for each environment: in Ring at density range $70$\textendash$200~veh/km$ for $50$ epochs with test accuracy of $95.5\%$ and in Bottleneck at inflow range $3000$\textendash$4500~veh/hr$ for $100$ epochs, achieving test accuracy of $85.2\%$.

Since the objectives, safety, stability, and efficiency, have intrinsic conflicts~\cite{zhu2020safe} (e.g., optimizing for throughput may lead to aggressive driving behaviors, thus compromising safety and stability), we propose two types of RVs: the \emph{safety + stability} RV, which prioritizes safety and stability, and the \emph{efficiency} RV, which emphasizes efficiency.   
Both types of RVs leverage CSC, and share the action and observation space in both Ring and Bottleneck. We train our RVs using the PPO algorithm~\cite{schulman2017proximal} with the following formulation:

\begin{itemize}[leftmargin=*]
    \item \textbf{Observation}. RV observes its leader vehicle's position and velocity, as well as the output of CSC: 
    $$\left\langle v_{n}, r_p(n, n-1), r_v(n, n-1) \right\rangle \oplus f_{\text{CSC}}\left(\{r_{p(i)}, r_{v(i)}\}_{i \in Z}\right), $$ 
    where $v_n$ is the velocity of the RV, $r_p(n,~n-1)$ and $r_v(n,~n-1)$ are the relative position and velocity respectively of RV with its immediate leader. For all $|Z|$ vehicles in the sensing zone (set to $50~m$), the relative position $r_{p(i)}$ and velocity $r_{v(i)}$ are input to CSC to predict the congestion stage $f_{\text{CSC}}(\cdot)$.
    \item \textbf{Action}. RV controls its acceleration within $[-3,3]~m/s^2$. 
    \item \textbf{Reward}.
Our reward is a linear combination of either RV velocity or average velocity of all vehicles, RV acceleration penalty, and a shaping component based on CSC output. 
For the two types of RVs, respective rewards are: 



\vspace{-8pt}
\begin{algorithm}[H]
\caption*{\small{Reward Functions}} 
\small{\texttt{Ring and Bottleneck: \textit{efficiency} RV}}
\begin{algorithmic}
\State{\small{$\text{reward} \leftarrow 0.75 \times v^* - 2\times |a_n|$}}
\If{\small{$f_{\text{CSC}}(\cdot)~=~\text{Congested~and~}$}\small{{$~\text{sign}(a_n)~>~0$}}}
\State{\small{$\text{reward} \leftarrow \text{reward} \plus \min(-1,~\lambda_1 \cdot|a_n|)$}} 
\EndIf{}
\If{\small{$f_{\text{CSC}}(\cdot)~=~\text{Leaving~and~}$}\small{{$~\text{sign}(a_n)~<~0$}}}
\State{\small{$\text{reward} \leftarrow \text{reward} \plus~\lambda_2 \cdot|a_n|$}} 
\EndIf{}
\end{algorithmic}
\vspace{2pt}
\small{\texttt{Ring: \textit{safety\plus stability} RV}}
\begin{algorithmic}
\State{\small{$\text{reward} \leftarrow 0.15 \times v^* - 4\times |a_n|$}}
\If{\small{$f_{\text{CSC}}(\cdot)~=~\text{Forming}$}}
\State{\small{$\text{reward} \leftarrow \text{reward} \plus \min(-1,~\lambda_3 \cdot|a_n|)$}} 
\EndIf{}
\end{algorithmic}
\vspace{2pt}
\small{\texttt{Bottleneck: \textit{safety\plus stability} RV}}
\begin{algorithmic}
\State{\small{$\text{reward} \leftarrow 0.5 \times v^* - 4\times |a_n|$}}
\If{\small{$f_{\text{CSC}}(\cdot)~=~\text{Forming~$|$~Congested~$|$~Undefined}$}}
\State{\small{$\text{reward} \leftarrow \text{reward} \plus \lambda_4 \cdot v_n$}} 
\If{\small{{$~\text{sign}(a_n)~\ge~0$}}}
\State{\small{$\text{reward} \leftarrow \text{reward} \plus \min(-2,~\lambda_5 \cdot|a_n|)$}} 
\EndIf{}
\EndIf{}
\end{algorithmic}
\end{algorithm}

\vspace{-10pt}
where $v^* = \frac{4}{3n} \sum_{i=1}^{n} v_i$ in Ring with $v_i$ as velocity of $i^{th}$ vehicle and $v^* = v_n$ in Bottleneck, $f_{\text{CSC}}(\cdot)$ is the CSC output, $a_n$ and $v_n$ are the acceleration and velocity of the RV, respectively. 
$\lambda_{1}=-10,~\lambda_{2}=-10,~\lambda_{3}=-5,~\lambda_{4}=-4$, and $\lambda_5 = -20$ are set empirically.

\item \textbf{Scaling laws}. In Ring, for RV penetration rates $>5\%$, both \textit{safety\plus stability} and \textit{efficiency} RVs are configured as platoons with a single leader and multiple followers (see Fig.~\ref{fig:systematic} MIDDLE). 
For example, our \textit{efficiency} RV platoon at $40\%$ penetration consists of $9$ RVs ($\round{22\times0.4}=9$) including a leader \textit{efficiency} RV trained at $5\%$ penetration rate and $8$ follower RVs. The follower RVs observe the position and velocity of the entire platoon and optimize the following reward: 
$$
    \small{\lambda_6\cdot r_p(n, n+j) + \lambda_7\cdot r_v(n, n+j) +\lambda_8\cdot |a_{n+j}| +\lambda_9}, 
$$
where $r_p(n, n+ j)$ and $r_v(n, n+ j)$ are relative position and relative velocity of the $j^\text{th}$ follower with the platoon leader, respectively; and $a_{n+ j}$ is the acceleration of the $j^{th}$ follower. $\lambda_6 = -2,~\lambda_7 = 4,~\lambda_8 = -4$,~and~$\lambda_9 = 10$ are chosen empirically. Whereas in Bottleneck, our RVs are dispersed and share the same RL policy.
\end{itemize}

We benchmark \model{} with two RL-based RVs. In Ring, we use RV(s) with only microscopic/local observations such as the relative position and velocity to its leader vehicle~\cite{wu2021flow}, referred to as RL\plus L hereafter. 
Whereas in  Bottleneck, we use RV(s) with macroscopic/global observations such as traffic density~\cite{vinitsky2018benchmarks}, referred to as RL\plus G hereafter.

\begin{table}[t]
\begin{center}
\vspace{6pt}
\normalsize
\setlength{\tabcolsep}{8pt} 
    \begin{tabular}{lccc} 
        \toprule
        \multirow{3}{*}{Controller}& Min. no. of & Time to & Average   \\ 
        & Stabilizing & Stabilize & Velocity\\
        & vehicles & (s) & (m/s)\\
        \toprule 

        IDM   & Unstable & Unstable & $3.58$ \\
        FS    & $1$ & $108$ & $5.20$ \\ 
        PIwS  & $1$ & $119$ & $5.33$ \\ 
        RL\plus L & $1$ & $74$ & $5.25$ \\
        Ours ($5$\%) & $1$ & $130$ & $5.28$ \\
        \hline 
        BCM   & $4$ & $146$ & $5.26$ \\ 
        Ours ($20$\%) & $4$ & $59$ & $4.86$ \\
        \hline 
        LACC  & $9$ & $690$ & $5.25$ \\ 
        Ours ($40$\%) & $9$ & $185$ & $5.53$ \\
        \bottomrule
    \end{tabular}
\end{center}
\vspace{-5pt}
\caption{\small{RVs' stabilization metrics in Ring at $81~veh/hr$ with HVs enabled by IDM. Stability is established when the standard deviation of the average velocity is below the IDM noise threshold $0.2$~\cite{chou2022lord}. Metrics are averaged over $10$ randomized simulation runs.}}
\vspace{-12pt}
\label{table:controllers}
\end{table}

\begin{figure*}[t!]
		\centering
		\includegraphics[width=0.98\linewidth]{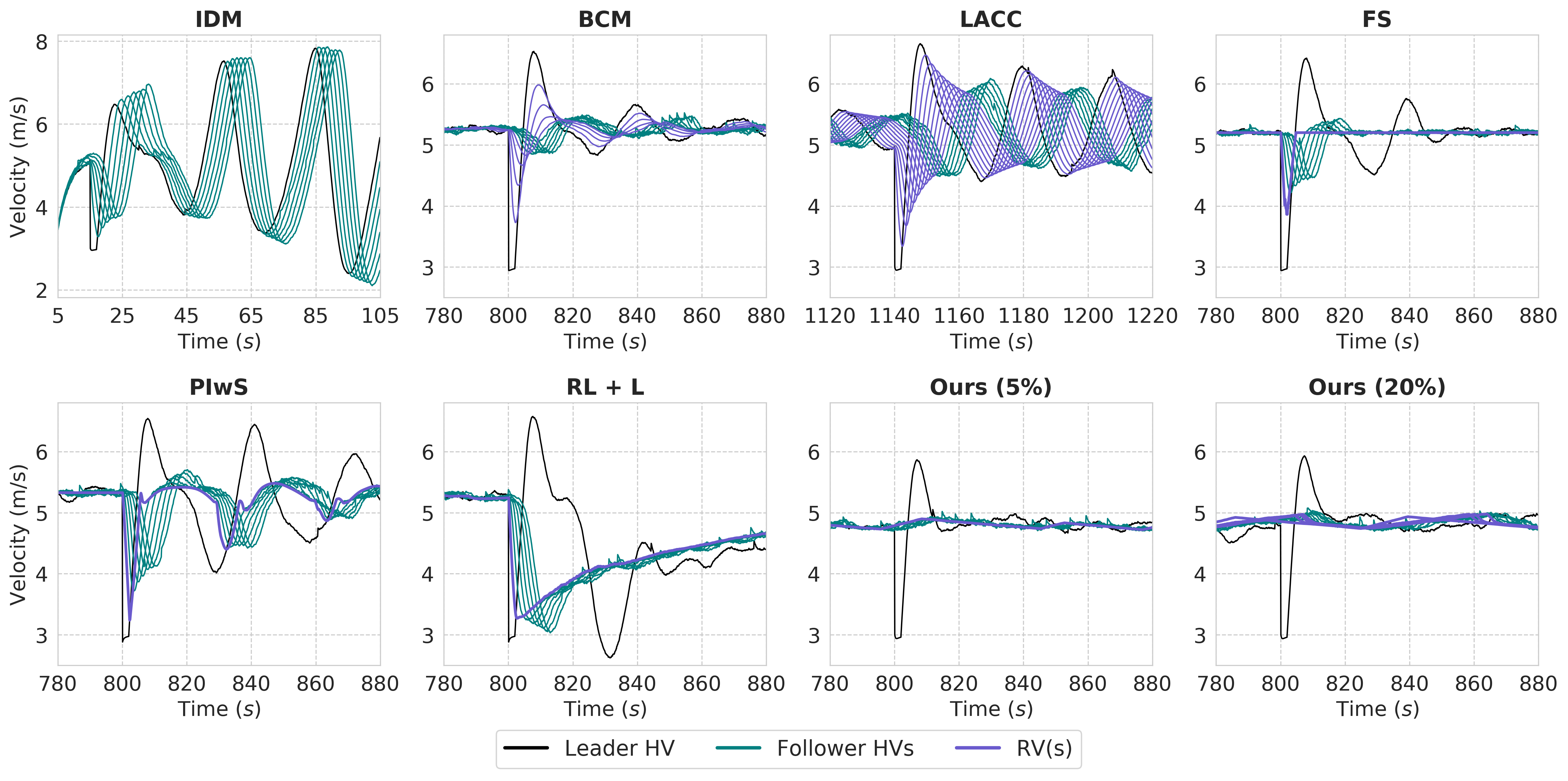}
        \vspace{-4pt}
		\caption{\small{Progressive amplification of longitudinal perturbation in the absence of RVs, and the strategies adopted by various RVs to attenuate such perturbations and stabilize traffic ($6$ HVs following the RVs are shown). Wave attenuation characteristics of various RVs at $81~veh/km$ are provided. The HV immediately in front of RV(s) produces a shock by applying a standard velocity perturbation of $3~m/s$. With IDM (100\% HVs), perturbation amplifies over time, whereas RVs dampen the perturbation over time.}}
		\label{fig:stability}
    \vspace{-12pt}
\end{figure*}
\section{Experiments}
We introduce the mixed traffic environments, the evaluation metrics, the experimental setup, and finally the results.
To begin with, we test on two mixed traffic environments.
\begin{itemize}[leftmargin=*]
    \item \textbf{Ring}: a single-lane circular road network with $22$ vehicles (see Fig.~\ref{fig:systematic} MIDDLE). This classical scene illustrates congestion development without external disturbances.
\item \textbf{Bottleneck}: a straight road where the number of lanes reduces from $8$ to $4$ and then to $2$. This simulates vehicles experiencing a capacity drop~\cite{saberi2013hysteresis} (see Fig.~\ref{fig:systematic} BOTTOM).   
\end{itemize}


\noindent Our evaluation metrics include the following.

\begin{itemize}[leftmargin=*]
    \item \textbf{Time to Collision (TTC)}: the time interval between two vehicles that will collide if they maintain a relative speed difference~\cite{vogel2003comparison}: 
$$
    TTC = 
    \begin{cases}
        &\frac{s - l}{v_f - v_l}, v_f>v_l, \\
        &\small{\infty}, v_f\leq v_l.
    \end{cases}
    \label{eq:ttc}
$$
$s$ is the space headway, $l$ is the vehicle length, and $v_f, v_l$ are the velocities of the RV and its HV leader, respectively. A lower TTC indicates a higher risk.

\item \textbf{Deceleration Rate to Avoid a Crash (DRAC)}: the force experienced by a vehicle in an emergent braking in order to avoid a front-end collision:
$$
    DRAC = 
    \begin{cases}
        & \frac{(v_f -v_l)^2}{s-l}, v_f>v_l, \\
        & 0, v_f\leq v_l.
    \end{cases}
    \label{eq:drac}
$$
$s$ is the space headway, $l$ is the vehicle length, and $v_f, v_l$ are the velocities of the RV and its HV leader, respectively. A lower DRAC represents a safer situation~\cite{cooper1976traffic}. 

\item \textbf{Fuel Economy (FE)}: the average fuel consumption of all vehicles in the network measured in miles per gallon ($mpg$) using the Handbook Emission Factors for Road Transport $3$ Euro $4$ passenger car emission model~\cite{de2004modelling}.

\item \textbf{Throughput}: the network flow rate of vehicles ($veh/hr$).

\item \textbf{Controller Acceleration Variation (CAV)}: the standard deviation of RV acceleration. A higher CAV meaning an RV is overly sensitive to its inputs, causing perturbations in upstream traffic~\cite{shi2022bilateral}.

\item \textbf{Wave Attenuation Ratio (WAR):} measures the effectiveness of an RV in dampening perturbations~\cite{tak2016study}. For all RVs, a standard velocity perturbation is applied to their immediate leader HV:
$$
    \text{WAR} = 1 - \frac{\Delta v_{\text{follow\_HV}}}{\Delta v_{\text{lead\_HV}}}.\\
    \label{eq:war}
$$
$\Delta v_{\text{follow\_HV}}$ is the velocity drop in the HV that immediately follows RV(s) and $\Delta v_{\text{lead\_HV}}$ is the velocity drop in the HV that immediately leads RV(s). A higher WAR indicates better dampening effect. 
\end{itemize}



To set up the experiments, we use FLOW~\cite{wu2021flow} and SUMO~\cite{lopez2018microscopic}.
Various vehicle configurations are adopted: RVs are platooned in Ring when penetration rate is $>5\%$ and dispersed in Bottleneck at all penetration rates. 
Both configurations have shown to stabilize traffic~\cite{chou2022lord}. We select penetration rates $5\%$, $20\%$, $40\%$, and $60\%$ to align with the minimum rates required for stabilizing traffic by different controllers as shown in Table~\ref{table:controllers}. To pursue rigorous evaluations of safety (TTC and DRAC) and CAV, when multiple RVs are present, we report the worst-case values of \model{}. For the IDM baseline, worst-case values of TTC and DRAC are again reported, whereas CAV is measured for a single randomly selected vehicle. For each experiment, an RV is first given sufficient time to stabilize traffic (as reported in Table~\ref{table:controllers}), followed by applications of acceleration perturbations for six minutes. 
Under these acceleration perturbations, we measure the safety of RVs.
Since only RVs are responsible for stabilizing the system, we also measure the stability of RVs. 
All vehicles in the network are accounted when measuring traffic efficiency. 



\begin{table*}[ht!]
\vspace{6pt}
\centering
\setlength{\tabcolsep}{5.75pt} 
\renewcommand{\arraystretch}{1.35} 
\begin{center}
\scalebox{1.0}{
    \begin{tabular}{|c|l|cc|cc|cc|cc|cc|cc|}
        \hline
         RV & & \multicolumn{6}{c|}{Ring} & \multicolumn{6}{c|}{Bottleneck}\\
        \cline{3-14}
         Pen. & RV Type & \multicolumn{2}{c|}{Safety} & \multicolumn{2}{c|}{Efficiency} & \multicolumn{2}{c|}{Stability}& \multicolumn{2}{c|}{Safety} & \multicolumn{2}{c|}{Efficiency} & \multicolumn{2}{c|}{Stability} \\ 
        \cline{3-14} 
          Rate & & TTC & DRAC & FE & Throughput & CAV & WAR & TTC & DRAC & FE & Throughput & CAV & WAR \\ 
        \hline
        & IDM$^{*}$ & $1.25$ & $1.19$ & $7.16$ & $986$ & $0.83$ & $\text{Unstable}$ & $1.23$ & $5.40$ & $15.84$ & $1652$ & $1.29$ & $\text{Unstable}$ \\
        \cline{1-14}
        \multirow{6}{*}{$5\%$} & FS & $3.52$ & $0.80$ & $11.35$ & $1214$ & $0.47$ & $0.54$ & $1.40$ & $5.07$ & $16.09$ & $1770$ & $1.53$ & $0.54$\\
        & PIwS & $1.24$ & $2.53$ & $10.87$ & $1260$ & $0.34$ & $0.61$ & $1.35$ & $5.00$ & $\bf{16.37}$ & $1686$ & $1.44$ & $0.61$\\
        & BCM & $1.03$ & $2.87$ & $7.58$ & $1046$ & $0.74$ & $\text{Unstable}$ & $1.51$ & $4.93$ & $16.06$ & $1642$ & $1.42$  & $0.26$\\
        & LACC & $1.08$ & $1.40$ & $7.26$ & $1036$ & $0.80$ & $\text{Unstable}$ & $1.44$ & $5.65$ & $15.31$ & $1594$ & $1.57$ & $0.12$\\\
        & RL\plus L/G& $2.25$ & $1.38$ & $9.28$ & $1060$ & $0.27$ & $0.47$ & $1.24$ & $4.74$ & $15.24$ & $1682$ & $1.43$ & $0.17$\\
        & Ours & $\bf{5.07}$ & $\bf{0.54}$ & $\bf{11.44}$ & $\bf{1272}$ & $\bf{0.02}$ & $\bf{0.91}$ & $\bf{4.70}$ & $\bf{2.42}$ & $10.90$ & $\bf{2086}$ & $\bf{0.85}$ & $\bf{0.91}$\\
        \cline{1-14}
        \multirow{6}{*}{$20\%$} & FS & $3.61$ & $0.71$ & $11.43$ & $1318$ & $0.39$ & $\bf{0.92}$ & $1.32$ & $5.81$ & $16.02$ & $1732$ & $1.51$ & $0.54$\\
        & PIwS & $1.65$ & $2.05$ & $10.51$ & $1296$ & $0.22$  & $\bf{0.92}$ & $1.25$ & $5.86$ & $\bf{16.69}$ & $1750$ & $1.61$ & $0.61$\\
        & BCM & $1.12$ & $1.49$ & $12.28$ & $1342$ & $0.41$ & $0.89$ & $1.36$ & $5.46$ & $16.32$ & $1666$ & $1.61$ & $0.26$\\
        & LACC & $1.09$ & $1.69$ & $8.19$ & $1148$ & $0.82$ & $\text{Unstable}$ & $1.26$ & $5.61$ & $15.35$ & $1704$ & $1.75$ & $0.12$\\
        & RL\plus L/G& $1.97$ & $1.49$ & $4.91$ & $654$ & $\bf{0.12}$ & $0.47$ & $1.42$ & $5.01$ & $16.45$ & $1730$ & $1.57$  & $0.17$\\
        & Ours & $\bf{5.38}$ & $\bf{0.53}$ & $\bf{11.99}$ & $\bf{1378}$ & $0.18$ & $0.91$ & $\bf{2.66}$ & $\bf{1.83}$ & $9.22$ & $\bf{2270}$ & $\bf{0.94}$ & $\bf{0.91}$\\
        \cline{1-14}
        \multirow{6}{*}{$40\%$} & FS & $2.86$ & $1.15$ & $10.79$ & $1346$ & $0.38$ & $\bf{0.97}$  & $1.26$ & $5.70$ & $17.16$ & $1754$ & $1.62$  & $0.54$\\
        & PIwS & $1.76$ & $1.75$ & $9.62$ & $1284$ & $0.21$  & $0.90$ & $1.25$ & $5.89$ & $16.86$ & $1750$ & $1.86$ & $0.61$\\
        & BCM & $1.96$ & $1.02$ & $10.93$ & $1428$ & $0.29$ & $\bf{0.97}$ & $1.27$ & $6.07$ & $16.69$ & $1726$ & $1.78$ & $0.26$\\
        & LACC & $4.42$ & $0.76$ & $12.35$ & $1444$ & $0.28$ & $0.72$ & $1.19$ & $5.48$ & $16.76$ & $1738$ & $1.85$ & $0.12$\\
        & RL\plus L/G& $1.92$ & $1.42$ & $2.58$ & $338$ & $0.37$ & $0.47$ & $1.22$ & $4.54$ & $\bf{17.54}$ & $1694$ & $1.44$ & $0.17$\\
        & Ours & $\bf{4.59}$ & $\bf{0.72}$ & $\bf{12.25}$ & $\bf{1448}$ & $\bf{0.18}$ & $0.91$ & $\bf{1.86}$ & $\bf{2.28}$ & $8.68$ & $\bf{2340}$ & $\bf{0.96}$ & $\bf{0.91}$\\ 
        \cline{1-14}
        \multirow{6}{*}{$60\%$} & FS & $2.50$ & $0.95$ & $8.86$ & $1128$ & $0.49$ & $\bf{0.97}$ &  $1.20$ & $5.66$ & $18.18$ & $1826$ & $1.96$  & $0.54$\\
        & PIwS & $1.60$ & $1.94$ & $9.67$ & $1324$ & $0.22$ & $0.95$ & $1.16$ & $6.11$ & $18.12$ & $1884$ & $1.96$ & $0.61$\\
        & BCM & $2.25$ & $0.58$ & $10.36$ & $1382$ & $0.23$ & $\bf{0.97}$ & $1.17$ & $6.15$ & $18.26$ & $1886$ & $1.95$ & $0.26$\\
        & LACC & $2.90$ & $0.85$ & $11.77$ & $1454$ & $0.28$ & $0.86$ & $1.33$ & $5.61$ & $18.63$ & $1868$ & $1.84$ & $0.12$\\
        & RL\plus L/G& $1.88$ & $1.57$ & $2.03$ & $280$ & $0.25$ & $0.47$ & $1.42$ & $4.41$ & $\bf{19.90}$ & $1924$ & $1.61$ & $0.17$\\
        & Ours & $\bf{4.42}$ & $\bf{0.81}$ & $\bf{12.58}$ & $\bf{1524}$ & $\bf{0.18}$ & $0.91$ & $\bf{1.55}$ & $\bf{2.57}$ & $7.82$ & $\bf{2034}$ & $\bf{1.04}$ & $\bf{0.91}$\\
        \hline
    \end{tabular}}
\end{center}
\vspace{-8pt}
\caption{\small{Evaluations of various RVs and our method. IDM* denotes $100\%$ HVs as baseline. LEFT: In Ring, under density $85~veh/km$, our \textit{safety\plus stability} RV uniquely achieves the highest safety measure (TCC $> 4~s$) and low oscillations (CAV $< 0.19~m/s^{2}$) at all penetration rates, with a constant WAR ($0.91$). In contrast, WAR compounds for FS, PIwS, BCM, and LACC as penetration rate increases with BCM and LACC stabilizing traffic only at higher penetration rates ($20\%\plus$ and $40\%\plus$, respectively). Our efficiency RV significantly improves the throughput (up to $54\%$) and fuel economy (up to $75\%$), while RL\plus L's efficiency declines as the number of RVs in traffic increases. RIGHT: In Bottleneck, under peak density $150~veh/km$, our \textit{safety\plus stability} RV maintains the highest WAR ($0.91$) at all penetration rates with the highest reduction in DRAC (up to $66\%$). Our \textit{efficiency} RV experiences a significant increase in throughput at all penetration rates (up to $41\%$). Whereas at lower penetrations ($5\%$ and $20\%$) PIwS improves fuel economy by $3\%$ and $5\%$, respectively; at higher penetration rates ($40\%$ and $60\%$),  RL\plus G offers the highest fuel economy improvements at $10\%$ and $25\%$, respectively.}}
\label{tab:results}
\vspace{-15pt}
\end{table*}

We next report results. Fig.~\ref{fig:stability} illustrates both the progressive amplification of longitudinal perturbation in the absence of RVs ($100\%$ IDM vehicles), and the strategies used by various RVs to attenuate such perturbations and stabilize traffic. 
We use Ring with density $81~veh/km$ to demonstrate. 
Worth noting, the phenomenon shown in Fig.~\ref{fig:stability} is environment-agnostic and can be replicated in any environment where longitudinal perturbation applies.
Specifically, at the $1140^{\text{th}}~s$ for LACC, $150^{\text{th}}~s$ for IDM, and $800^{\text{th}}~s$ for the others, an identical velocity perturbation is applied to the HV that immediately leads the RV(s) (a randomly chosen HV in case of IDM), in which the velocity of the leader HV is abruptly decreased to $3~m/s$ for $2~s$. 
For IDM, the perturbation is introduced before stop-and-go waves, while for the others, it is applied after stop-and-go waves, and the RV(s) are given sufficient time to stabilize the traffic. We observe that in the absence of RVs, IDM vehicles oscillate within the range $[2, 8]~m/s$, periodically increasing and decreasing their velocity, and eventually forming a stop-and-go wave. 
In contrast, RVs attenuate the standard perturbation of the leader HV and stabilize traffic by adopting various strategies. For example, PIwS drives at an estimated average velocity and BCM maintains equal distance to its leader and follower HVs.   
The attenuation characteristics of our \textit{safety\plus stability} RVs are shown in Fig.~\ref{fig:stability} at $5\%$ and $20\%$ penetration rates denoted by Ours~$(5\%)$ and Ours~$(20\%)$ accordingly. 
Compared to RL\plus L, our RVs open up a wider gap in front and track their HV leader at a slightly lower velocity allowing them to dampen incoming perturbations. For Ours~$(20\%)$, multiple follower RVs closely track the leader RV. In Table~\ref{tab:results}, the results of the \textit{safety\plus stability} RV are shown in the Safety and Stability columns, while the results of the \textit{efficiency} RV are presented in the Efficiency column. All values in Table~\ref{tab:results} are averaged over $5$ randomized simulation runs.

Table~\ref{tab:results} LEFT presents the results of RVs in Ring at $85~veh/km$ density. Across all penetration rates, our \textit{safety\plus stability} RV delivers the best safety measures. 
Notably, only our RV manages to exceed the critical $4~s$ threshold for TTC, as recommended by earlier studies~\cite{ayres2001preferred, vogel2003comparison}. 
Additionally, DRAC is reduced by $54\%$ at $5\%$ penetration rate in comparison to IDM, and the CAV stays below $0.19~m/s^2$ across all penetration rates. 
This is because the gap opened by our \textit{safety\plus stability} RV dampens the perturbations caused by real-world disturbances. 
Furthermore, at $5\%,~20\%,~40\%,$ and $60\%$ penetration rates, our \textit{efficiency} RV improves the throughput by $29\%,~40\%,~46\%,$ and $54\%$, respectively, compared to IDM. 

Table~\ref{tab:results} RIGHT presents the results of RVs in Bottleneck at $150~veh/km$ peak density with an inflow rate of $3600~veh/hr$. Unlike Ring, where multiple RVs form a platoon, compounding the individual dampening effects, in Bottleneck, RVs are dispersed, with each RV aiming to dampen independently. 
Hence, the WAR for RVs remain constant at all penetration rates, among which our \textit{safety \plus stability} RV maintains the highest WAR at $0.91$, dampening majority of perturbations. 
Despite the high density in Bottleneck, our \textit{safety \plus stability} RV exceeds the critical $4~s$ TTC threshold at $5\%$ penetration rate, however, at higher penetration rates, TTC value is lower (but still the highest) among all other RVs. 
This is achieved by adopting the strategy of maintaining a front gap, consistent with its behavior in Ring.
At all penetration rates, our \textit{safety \plus stability} RV maintains the lowest DRAC (reduction of up to $66\%$ compared to IDM). 

Table~\ref{tab:results} RIGHT also shows that our \textit{efficiency} RV delivers the highest throughput at all penetration rates, up to $41\%$ higher than IDM at $40\%$ penetration rate. 
Since CSC does not encounter other RVs within the sensing zone during training, its accuracy decreases as more RVs appear in Bottleneck at higher penetration rates. This decline in performance is due to RVs behaving differently (usually lower accelerations and higher time gaps for \textit{safety \plus stability} RV and higher accelerations and lower time gaps for \textit{efficiency} RV) than IDM vehicles. 
Hence, CSC fails to anticipate the interactions between RVs and HVs within the sensing zone.
Furthermore, zipper lanes (tapered sections that merge $8\textendash4$ and $4\textendash2$ lanes) break any achieved stability, e.g., a string of vehicles stabilized by an RV, when reaches the zipper lane, often brakes to allow vehicles from adjacent lanes to safely merge and cut--in--front.
Lastly, the strategy adopted by our \textit{efficiency} RV favors throughput over fuel economy as our RVs are not optimizing fuel economy via the reward function.

\section{Conclusion and Discussion}

In this project, we incorporate real-world driving profiles into two mixed traffic environments, Ring and Bottleneck, with the goal to enhance the safety, efficiency, and stability of mixed traffic through RVs. For this purpose, we develop \model{} with two classes of RVs: \textit{safety\plus stability} RV and \textit{efficiency} RV. Both types of RVs leverage the congestion stage classifier to optimize their objectives. 
In Ring, our RVs are able to increase the time to collision (TTC) above the critical threshold of $4~s$, reduce the deceleration rate avoid a crash (DRAC) up to $54\%$, and increase the throughput up to $54\%$. This is achieved by dampening nearly all perturbations and maintaining the acceleration variation as low as $0.02~m/s^2$. 
In Bottleneck, our RVs are able to maintain the highest safety measure, i.e., TTC$>4~s$ at $5\%$ penetration rate and DRAC$<2.42$ at all penetration rates; and the highest throughput, i.e., improvements by up to $41\%$. 
Importantly, our RVs demonstrate a capacity to generalize and enhance performance when encountering more intricate dynamics, such as merges and cut-ins, within the zipper lanes.

While we acknowledge that passenger comfort alongside theoretical analyses such as stabilizability, controllability, and reachability~\cite{zheng2020smoothing, wang2020controllability} are significant and provide valuable insights, they fall beyond the scope of this work. 
We focus on directly measurable metrics, including enhancements in stability and reductions in oscillations. These not only offer quantitative assessments of performance but also directly correlate with other desired properties such as comfort.
Given communicating global traffic states using vehicle-to-vehicle or vehicle-to-infrastructure techniques is in nascent stages and may require costly upgrades to current traffic infrastructure, our approach offers a practical solution because it solely relies on observations of individual RVs. 
The limitations of our work include not using additional traffic features such as lane-changing and heterogeneous vehicle types, and assuming perfect sensing without temporal delays. 
These issues serve as interesting future directions that we plan to explore. 



\bibliographystyle{unsrt}
\bibliography{references}


\section*{APPENDIX} 
\section*{I. Car Following Filter} 
\begin{figure}[h!]
		\centering
        \vspace{2pt}
		\includegraphics[width=0.88\linewidth]{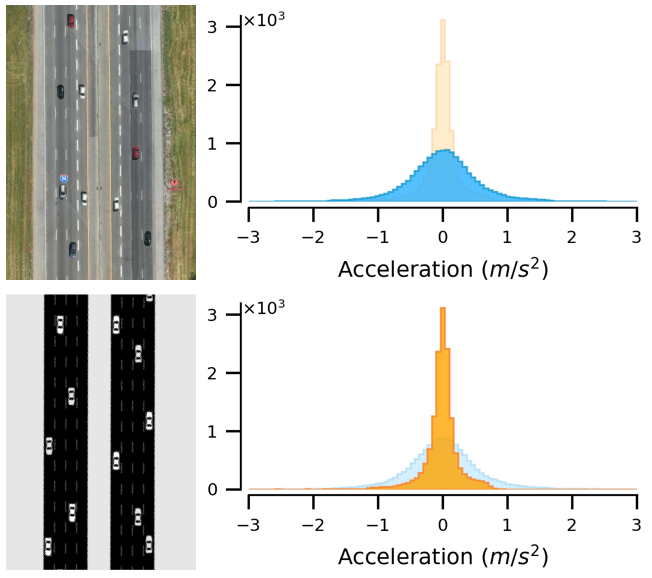}
        \vspace{1pt}
		\caption{\small{Instantaneous accelerations observed during car-following behaviors at densities $[70, 150]~veh/km$. TOP: Real-world data from the I-$24$ MOTION dataset reveals a distribution having long tails extending to $[-3, 3]~m/s^2$. BOTTOM: IDM (in simulation) produces accelerations mostly within $[-0.5, 0.5]~m/s^2$, indicating much `timid' driving behaviors than the real world.}}
		\label{fig:real_world}
    \vspace{-8pt}
\end{figure}
We analyze the I-$24$ MOTION dataset~\cite{gloudemans202324} with study length $=6.75~km$ and study time $=4~h$. The dataset contains various vehicle types such as semi-trailers, mid-sized trucks, motorbikes, and cars under different traffic conditions such as approaching standing traffic, lane changing, and free flow. 
To extract car-following trajectories, we select data points that meet the following criteria:
\begin{itemize} 
    \item Ego car is following another car, i.e., has a leader.
    \item Leader and ego cars are in the same lane $\ge 5~s$.
    \item Ego car's speed is greater than $10\%$ of the speed limit, i.e., not approaching stationary traffic. 
    \item Ego car's space headway is less than $124~m$, applying $4~s$ rule at the speed limit to avoid free flow conditions.
\end{itemize}

\section*{II. Model-based Robot Vehicles}
\label{subsec:controllers}
\textbf{Bilateral Control Module (BCM)}: BCM~\cite{horn2013suppressing} uses information about both follower and leader vehicles to obtain a linear model whose acceleration is given by: 
\vspace{-.5pt}
\begin{equation}
    a = k_d\cdot \Delta_d + k_v \cdot(\Delta v_l - \Delta v_f) + k_c \cdot (v_{des} - v), 
    \label{eq:bcm}
\end{equation}
where $\Delta_d$, $\Delta v_l$, $\Delta v_f$, $v_{des}$, and $v$, represent 
the difference in distance to the leader compared to the distance to the follower,
the difference in velocity to the leader, the difference in velocity to the follower, the set desired velocity, and the current velocity of the vehicle, respectively. $k_d=1$, $k_v=1$, and $k_c=1$ are gain parameters. 

\textbf{Linear Adaptive Cruise Control (LACC)}: LACC is an improvement on existing cruise control systems that allows vehicles to maintain a safe distance or speed without communication. The constant time-headway model by Rajamani~\cite{rajamani2011vehicle} employs a first-order differential equation for approximation. 
The control acceleration at time $t$ is given by:
\begin{flalign}
    a_{t} &= (1 - \frac{\Delta t}{\tau})\cdot a_{(t-1)} + \frac{\Delta t}{\tau}a_{cmd, (t-1)}, \label{eq:lacc_1} \\[-0.1em]
    a_{cmd} &= k_1\cdot e_x + k_2 \cdot \Delta v_l, \label{eq:lacc_2} \\[-0.1em]
    e_x &= s - h\cdot v, \label{eq:lacc_3}
\end{flalign}
where $k_1=0.3$ and $k_2=0.4$ are design parameters, $e_x$ is the gap error, $s$ is the space headway, $\Delta v_l$ is the velocity difference to the leader, $h=1$ is the desired time gap, $\Delta t$ is the control time-step, and $\tau=0.1$ is the time lag of the system.

\vspace{2pt}
\section*{III. Heuristic-based Robot Vehicles}
\textbf{FollowerStopper (FS)}: FS~\cite{stern2018dissipation} is an RV that travels at a fixed command velocity (target) under safe conditions but when required, slightly lowers the target velocity, opening up a gap to the vehicle ahead. This allows it to dampen oscillations and brake smoothly when needed.  The command velocity is given by:

\begin{equation}
v_{cmd} =
    \begin{cases}
    0, & \text{if } \Delta x \leq \Delta x_1 \\
    v \frac{\Delta x - \Delta x_1}{\Delta x_2 - \Delta x_1}, & \text{if } \Delta x_1 < \Delta x \leq \Delta x_2 \\
    v + (U - v) \frac{\Delta x - \Delta x_2}{\Delta x_3 - \Delta x_2}, & \text{if } \Delta x_2 < \Delta x \leq \Delta x_3 \\
    U, & \text{if } \Delta x_3 < \Delta x
    \end{cases}
    \label{eq:follower-stopper}
\end{equation}

\noindent where $v = \min\left(\max\left(v_{\text{lead}}, 0\right), U\right)$ is the speed of the leader vehicle, $\Delta x$ is the headway of the RV, and $U$ is the desired velocity.
The thresholds ($\Delta x_1,~\Delta x_2,~\Delta x_3$) are defined as
\begin{equation}
\Delta x_k = \Delta x_k^{0} + \frac{1}{2d_k}(\Delta v_{-})^2, \quad k = 1, 2, 3. 
\label{eq:follower-stopper-boundary}
\end{equation}
The model parameters $\Delta x_k^{0}$, $\Delta v_{-}$, and $d_k$ determine the spacing between vehicles and the RV's responsiveness to changes in velocity.

\textbf{Proportional-integral with saturation (PIwS)}: PIwS~\cite{stern2018dissipation} estimates the desired average velocity ($U$) of the vehicles in the network using its historical average velocity. The PIwS RV calculates the target velocity as
\begin{equation}
    v_{target} = U + v_{catch} \times \min \left( \max \left( \frac{\Delta x - g_l}{g_u-g_l}, 0 \right), 1 \right), 
\label{eq:piws_1}
\end{equation}
which is used to calculate the command velocity at $t+1$ as 
\begin{flalign}
v_{cmd}^{t+1} = \beta_{t} (\alpha_{t} v_{target}^{t} + (1 - \alpha_{t}) v_{lead}^{t}) + (1 - \beta_{t}) v_{cmd}^{t}, 
\label{eq:piws_2}
\end{flalign}
where $v_{catch}$ is the catch-up velocity---a velocity higher than the average velocity allows the RV to catch up with its leader, $\Delta x$ is the difference in position between the RV and its leader, $g_l$ and $g_u$ represent the lower and upper threshold distance, respectively; $\alpha_{t}$ and $\beta_{t}$ represent the weight factors for target velocity $v_{target}$ and command velocity $v_{cmd}$, respectively. Finally, $v_{lead}$ represents the velocity of the leader vehicle.



\begin{figure*}[h!]
    \centering
    \centering
    \includegraphics[width=0.95\linewidth]{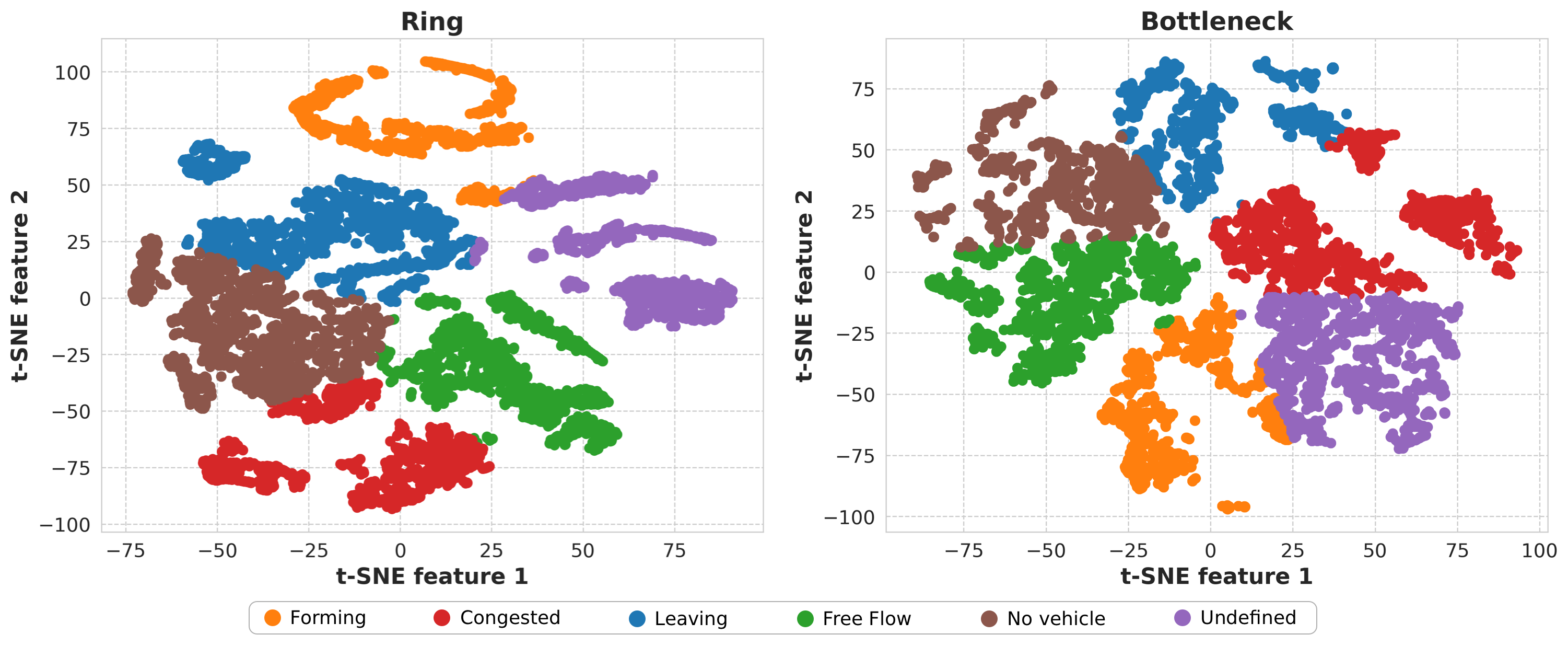}
    \vspace{-2pt}
    \caption{\small{The results of applying K-means clustering with t-SNE on a subset of CSC training data. LEFT: In Ring, the clusters are spread out, suggesting that the data is easily classifiable. RIGHT: In Bottleneck, overlapping clusters indicate that more complex interactions exist among the congestion stages, possibly due to the presence of zipper lanes causing vehicles abruptly merge.}}
    \label{fig:kmeans}
\end{figure*}

\begin{figure*}[h!]
    \centering
    \hfill
    \begin{minipage}{0.465\textwidth}
        \centering
        \includegraphics[width=\linewidth]{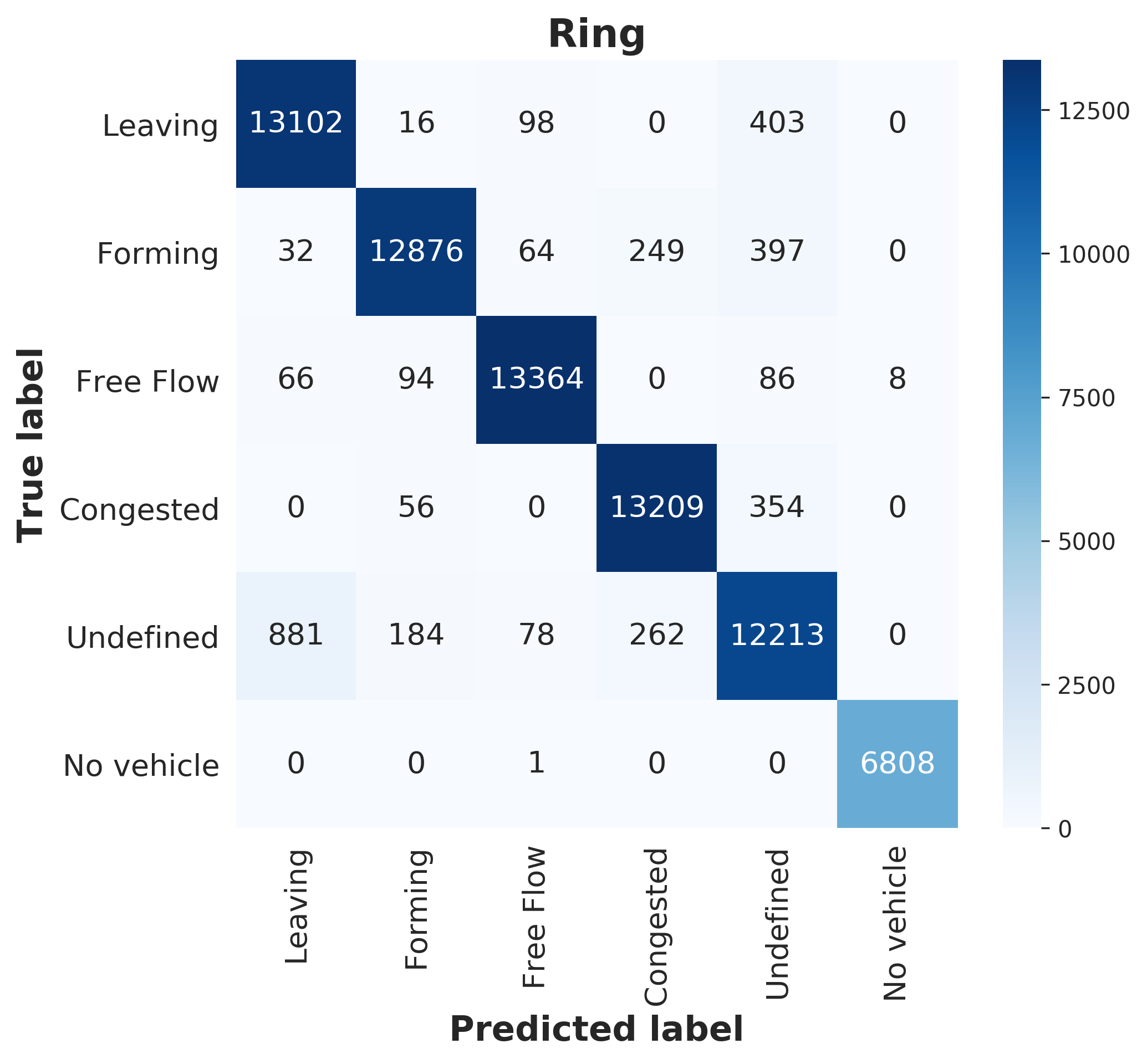}
    \end{minipage}
    \begin{minipage}{0.465\textwidth}
        \centering
        \includegraphics[width=\linewidth]{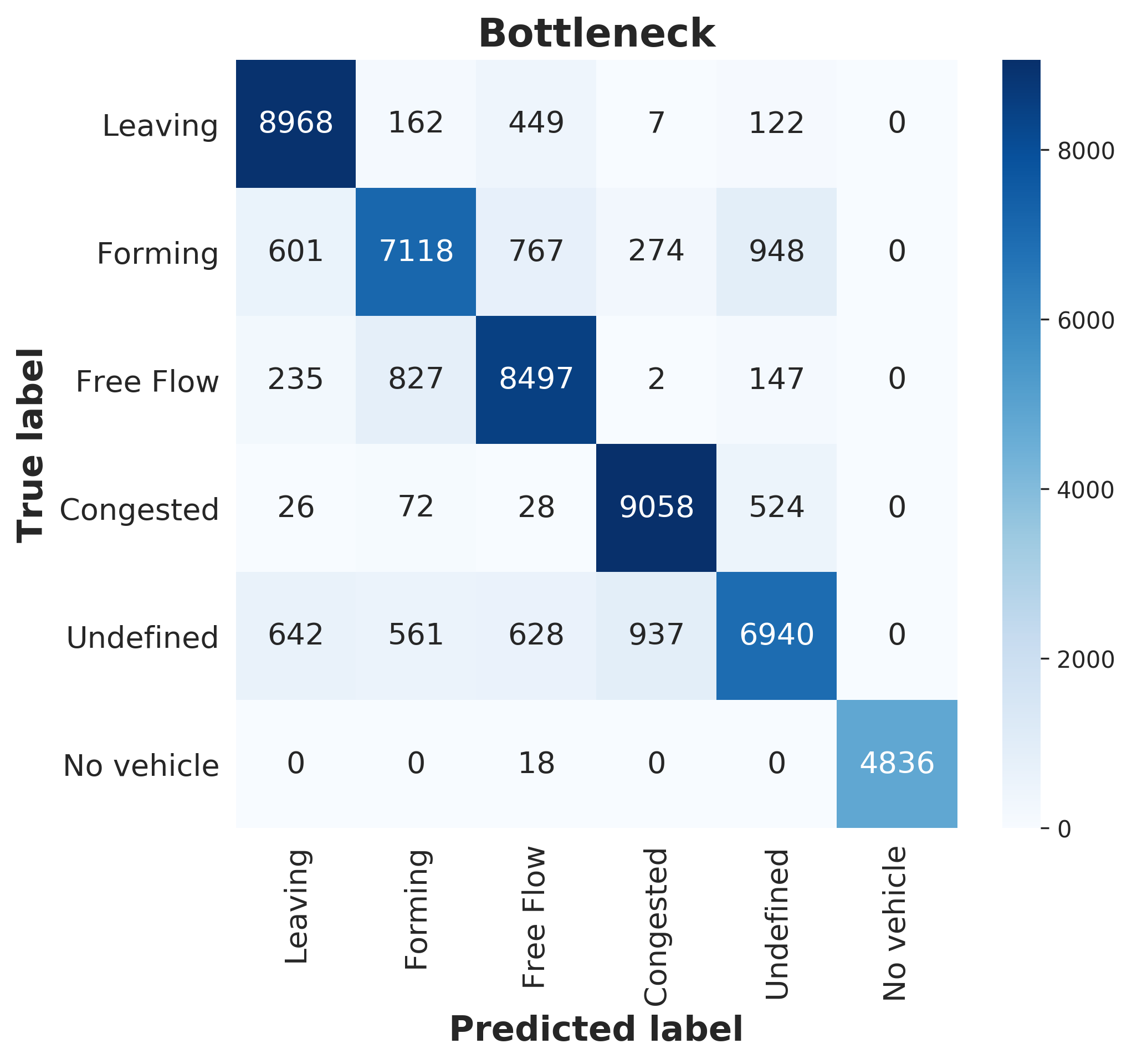}
    \end{minipage}
    \hfill
    \vspace{-2pt}
    \caption{\small{Confusion matrix of a trained CSC in Ring (LEFT) and Bottleneck (RIGHT) on the validation set.}}
    \vspace{-12pt}
\label{fig:confusion}
\end{figure*}

\begin{table}[h!]
\begin{center}
  \normalsize
  \setlength{\tabcolsep}{6pt}
  \begin{tabular}{llr}
    \toprule
    Category & Parameter& Value \\
    \toprule    
    \multirow{5}{*}{Ring} 
    & Time Step ($\Delta t$)& $0.1$ \\
    & Simulation Horizon ($T$) & $4500$ \\
    & Warmup Time-steps & $2500$ \\
    Simulation& Speed Limit ($m/s$) & $30$\\
    &Initial Speed ($m/s$) & $0$\\
    \hline    
    \multirow{5}{*}{Bottleneck} 
    & Time Step ($\Delta t$)& $0.5$ \\
    & Simulation Horizon ($T$) & $1300$ \\
    & Warmup Time-steps & $100$ \\
    Simulation& Speed Limit ($m/s$) & $17$\\
    & Initial Speed ($m/s$) & $6$\\
    & Inflow Rate ($veh/hr$)& $3600$\\
    \hline
    \multirow{7}{*}{PPO} 
    & Learning Rate ($\alpha$) & $0.00005$ \\
    & Discount Factor ($\gamma$) & $0.999$\\
    & GAE Estimation ($\lambda$) & $0.97$ \\
    & KL Divergence Target & $0.02$ \\
    Algorithm& Entropy Coefficient Initial& $0.1$\\
    & Entropy Coefficient Final& $0.01$\\
    & Value Function Clip Param& $20$\\
    & SGD Iterations& $2$\\
    \hline
    \multirow{3}{*}{Ring} 
    & Neural Network& $32, 16, 16$ \\
    & Batch Size& $32$ \\
    CSC& Learning Rate& $0.01$\\
    & Epochs& $50$\\
    \hline
    \multirow{3}{*}{Bottleneck} 
    & Neural Network& $32, 16, 16$ \\
    & Batch Size& $256$ \\
    CSC & Learning Rate& $0.001$\\
    & Epochs& $100$\\
    \hline
    \multirow{5}{*}{Policy} 
    & Our leader RV in Ring& $64, 32, 16$ \\
    & Our follower RV in Ring& $64, 32, 16$ \\
    & Our RV in Bottleneck& $32, 16, 8$ \\
    Networks & RL\plus L& $32, 32, 32$\\
    & RL\plus G& $256, 256$\\
    \bottomrule
  \end{tabular}
\end{center}
\vspace{-6pt}
\caption{\small{Detailed experiment parameters. We show the simulation parameters of Ring and Bottleneck, as well as the parameters of Proximal Policy Optimization (PPO) and Congestion Stage Classifier (CSC). The hidden layer dimensions of various policy networks are also shown.}}
\vspace{-16pt}
\label{table:params}
\end{table}

\vspace{2pt}
\section*{IV. Reinforcement Learning (RL) Benchmarks} 
To benchmark with other RL techniques, we reproduce their original policies by following the provided experiment parameters and closely matching the performance. 
Specifically, to obtain RL policy with only local observations (RL\plus L), we follow Wu et al.~\cite{wu2021flow}; to obtain RL with global observations (RL\plus G), we follow Vinitsky et al.~\cite{vinitsky2018lagrangian}. 
Our reproduced RL\plus L achieves the performance within $1\%$ error (measured with stabilization time and average velocity during stabilization) of the original work. Whereas for RL\plus G, our reproduction achieves the performance within $3\%$ error (measured with outflow) of the original work. 
The precise implementations of the other RL methods validate our benchmarking experiments.



\section*{V. Congestion Stage Classifier (CSC)} 
One CSC each is trained in Ring and Bottleneck with independent datasets collected in the two environments. 
For each RV, the position and velocity of all vehicles in its local zone (set to $50~m$) are collected. 
Fig.~\ref{fig:kmeans} shows the K-means clustering of collected data over all six classes (`Forming', `Leaving', `Congested', `Free flow', `Undefined', and `No Vehicle') in both environments. 
As the data collected is sequential in nature and CSC predictions are made a number of time-steps into the future, a time offset of $10$ time-steps is chosen to balance usefulness and accuracy (to illustrate, a prediction of congestion stage $100$ time-steps in the future would be very useful, however, not very accurate; whereas a prediction of congestion state $1$ time-step into the future can be very accurate but not very useful). 

After windowing, the dataset includes instances where the congestion stage changes from \textit{t} to \textit{t$\plus10$}, as well as instances where the congestion stage remains the same over the time window. To train CSC, we sample data to ensure a balanced representation of transition/non-transition instances as well as instances containing all six classes. Worth noting, the `No vehicle' class presents a unique challenge. The collected data may contain instances changing from `No vehicle' to another class after the $10$ time-steps. 
However, based on the input corresponding to `No vehicle' at \textit{t}, we cannot predict the congestion stage at \textit{t$\plus10$}. 
Consequently, we discard data points where the `No Vehicle' class transitions to another class after $10$ time-steps. We replace this discarded data with synthetic examples that simulate various scenarios for the RV's position and velocity without leader vehicles. 

The accuracy of the trained CSC is $95.5\%$ in Ring and $85.2\%$ in Bottleneck. The confusion matrix is shown in Fig.~\ref{fig:confusion}.  
CSC only observes downstream HVs in the same lane. 
Thus, when facing zipper lanes of Bottleneck (where traffic merges from adjacent lanes), the CSC cannot anticipate the merging traffic, resulting in lower accuracy in  Bottleneck. 
The CSC training parameters are provided in Table~\ref{table:params}.

\end{document}